%% file: main.tex
\documentclass[10pt,twocolumn,letterpaper]{article}

\usepackage{cvpr}              

\input{preamble}

\definecolor{cvprblue}{rgb}{0.21,0.49,0.74}
\usepackage[pagebackref,breaklinks,colorlinks,citecolor=cvprblue]{hyperref}

\def\paperID{*****} 
\def\confName{CVPR}
\def\confYear{2024}

\title{Learning Multimodal Latent Space with EBM Prior and MCMC Inference}

\author{$\text{Shiyu Yuan}^1$, $\text{Carlo Lipizzi}^1$, $\text{Tian Han}^2$\\
School of Systems and $\text{Enterprises}^1$, Department of Computer $\text{Science}^2$\\
Stevens Institute of Technology\\
{\tt\small \{syuan14, clipizzi, than6\}@stevens.edu}
}

\begin{document}
\maketitle
\input{sec/0_abstract}
\input{sec/1_intro}

\input{sec/2_relatedwork}
\input{sec/3_method_v3}
\input{sec/4_exp}
\input{sec/5_ablation}
\input{sec/5_conclusion}

{
    \small
    \bibliographystyle{ieeenat_fullname}
    \bibliography{main}
}


\end{document}

%% file: preamble.tex
\usepackage[dvipsnames]{xcolor}


%% file: sec/0_abstract.tex
\begin{abstract}
Multimodal generative models are crucial for various applications. We propose an approach that combines an expressive energy-based model (EBM) prior with Markov Chain Monte Carlo (MCMC) inference in the latent space for multimodal generation. The EBM prior acts as an informative guide, while MCMC inference, specifically through short-run Langevin dynamics, brings the posterior distribution closer to its true form. This method not only provides an expressive prior to better capture the complexity of multimodality but also improves the learning of shared latent variables for more coherent generation across modalities. Our proposed method is supported by empirical experiments, underscoring the effectiveness of our EBM prior with MCMC inference in enhancing cross-modal and joint generative tasks in multimodal contexts.
\end{abstract}

%% file: sec/1_intro.tex
\vspace{-5mm}
\section{Introduction}
\label{sec:intro}
Multimodal generative models are important because of their ability to interpret, integrate, and synthesize information from diverse inputs. In these models, shared latent variables play a crucial role in integrating features from diverse modalities into a unified and informative representation for downstream generative tasks.
Recent works have explored multimodal generation through denoising-based networks \cite{ho2020denoising, ramesh2022hierarchical, bao2023one} or by learning representations via modality alignment, as seen in \cite{radford2021learning}. However, the former approaches often lack a shared representation of different modalities, while the latter may not support generation tasks

Variational Autoencoder (VAE)-based models \cite{kingma2013auto} can achieve both objectives: learning a shared representation through a latent aggregation mechanism \cite{wu2018multimodal, shi2019variational} and generating data using top-down generators. However, this approach inherits the limitations of traditional VAE models, notably the reliance on uni-modal priors that are not informative enough to capture the complexity of multimodality.

To tackle the problem of non-informative prior, we propose a joint training scheme for multimodal generation that employs an EBM prior with MCMC inference. This approach leverages an expressive prior to better capture multimodal data complexity. Additionally, the use of MCMC inference with Langevin dynamics improves the learning process of EBM. In summary, our contributions are as follows:
\begin{enumerate}
\item We propose the use of an EBM prior to replace the uni-modal prior in multimodal generation, enhancing the capture of multimodal data complexity.
\item We employ MCMC inference to more accurately approximate the true posterior, as compared with variational inference, which improve EBM learning.
\item We conduct empirical experiments on multimodal datasets to validate our proposed EBM prior with MCMC inference, demonstrating improvements in the multimodal generative model both visually and numerically.
\end{enumerate}

%% file: sec/2_relatedwork.tex
\section{Related Work}
\subsection{Multimodal Generative Models}
In the learning of multimodal generative models, two fundamental challenges arise: one is obtaining a shared representation that captures the common knowledge among modalities, and the other is cross-modal generation, which involves translating between modalities \cite{suzuki2022survey}. VAE-based multimodal generative models \cite{wu2018multimodal, shi2019variational, sutter2020multimodal, sutter2021generalized, hwang2021multi, palumbo2023mmvae+} have achieved good performance in learning such shared information and performing cross-modal generation, but they still face the non-informative prior limitation.

\subsection{Expressive Prior}
Due to the complexity of data distributions, recent works seek to utilize expressive priors to represent prior knowledge in generative models, such as hierarchical priors \cite{vahdat2020nvae, cui2023learning}, flow-based priors \cite{xie2023tale}, and energy-based priors \cite{pang2020learning}. However, such expressive priors are rarely discussed in the context of multimodal generation. 

\subsection{MCMC-based Inference}
MCMC inference enables sampling from distributions that are otherwise challenging to track directly. Several works present promising performance on generative tasks through MCMC inference, such as dual-MCMC teaching, alternating back-propagation and short-run MCMC as seen in \cite{han2017alternating, nijkamp2019learning, cui2023nips}. However, these methods are rarely used in multimodal generative modeling.




%% file: sec/3_method_v3.tex
\section{Methodology}
\subsection{Preliminaries}
\noindent \textbf{Multimodal Generative Model}
Multimodal generative models aim to learn the joint distribution of multimodal data. Suppose there are $M$ modalities; data in each modality is denoted as $x^{m}$, the entire dataset is denoted as $X = (x^{1}, x^{2} \cdots x^{m})$, and the shared latent variable is denoted as $z$. The joint probability $p(z,X)$ can be factorized into Eqn. \ref{eqn:joint_fact}. 
{
\small
\begin{equation}
    \begin{aligned}\label{eqn:joint_fact}
        p(z,X) = p(z)\prod p(x^{1}|z)p(x^{2}|z) \cdots p(x^{m}|z)
    \end{aligned}
\end{equation}
}

{\noindent Most multimodal models learn $p(z,X)$ through a shared latent variable. In multimodal generative models, VAE-based models can learn such shared latents through two foundation aggregation approach: POE \cite{wu2018multimodal} and MOE \cite{shi2019variational}, with MOE being the more commonly adopted one  \cite{sutter2020multimodal,sutter2021generalized,hwang2021multi,palumbo2023mmvae+}. MOE averages latent variables from each modality, given by $q_{\Phi}(z|X) = \frac{1}{M} \sum_{m=1}^M q_{\phi_{m}}(z|x^{m})$. Learning such models typically adopting ELBO as in traditional VAE models, as shown in Eqn.\ref{eqn:obj_moe}.}
{
\footnotesize
\begin{equation}
    \begin{aligned}\label{eqn:obj_moe}
    &L_\text{MOE}({\theta}) =  \frac{1}{M} \sum_{m=1}^{M} \left[E_{q_{\phi_{m}}(z|x^{m})} \log \frac{p_{\beta}(z, X)}{q_{\Phi}(z|X)} \right]\\
    &= \frac{1}{M} \sum_{m=1}^{M} \left[E_{q_{\phi_{m}}(z|x^{m})} \log p_{\beta_{m}}(x^{m}|z) \right.\\
    &\left.+ \sum_{\substack{n=1 \\ n \neq m}}^{M}E_{q_{\phi_{m}}(z|x^{m})} \log p_{\beta_{n}}(x^{n}|z) \right.\\
    &\left.- \text{KL}\left[q_{\Phi}(z|X) \parallel p(z)\right] \right]\\
    &= E_{q_{\Phi}(z|X)} \log p_{\beta}(X|z) - \text{KL}\left[q_{\Phi}(z|X) \parallel p(z)\right]
    \end{aligned}
\end{equation}
}

{\noindent Where $\theta$ comprises $(\beta, \phi)$, ${\beta}$ and ${\phi}$ denote the generator and inference model parameters, respectively. One limitation is that the objective includes a uni-modal prior $p(z)$  which cannot sufficiently capture the complexity of multimodal data space. In this work, we propose using an expressive EBM prior $p_{\alpha}(z)$ to replace the non-informative prior.} \\
\noindent \textbf{EBM on Latent Space}
Latent space EBM aims to learn a latent distribution with high probability by assigning low energy, as shown in Eqn.\ref{eqn:model_ebm}
{
\small
\begin{equation}
    \begin{aligned}\label{eqn:model_ebm}
    &p_{\alpha}(z) = \frac{1}{Z(\alpha)}\exp [f_{\alpha}(z)] \cdot p(z)
    \end{aligned}
\end{equation}
}

{\noindent Where $-f_{\alpha}(z)$ is the energy function, $p(z)$ is uni-modal distribution as EBM prior initialization, $Z(\alpha) = \int_z \exp [f_{\alpha}(z)] \cdot p(z) dz$ is the normalization term normally intractable. Learning an EBM prior in the latent space has shown promising performance on generative tasks \cite{pang2020learning, han2020joint, cui2023learning, zhang2021learning}, but the application of EBM prior in multimodal generative models is under-explored. Moreover, EBM's un-normalized exponential distribution as in Eqn. \ref{eqn:model_ebm} provides high flexibility in modeling the latent space and enhances its expressiveness in representing the complexity of multimodal data.}

\subsection{Method}
Due to the non-informative prior in Eqn.\ref{eqn:obj_moe} and the expressiveness of the EBM prior, we propose a model that follows the MOE aggregation framework and is jointly learned with EBM prior using MLE. The objective of the joint learning model MOE-EBM is as follows in Eqn. \ref{eqn:model_moebm}.
{\footnotesize
\begin{equation}
    \begin{aligned}\label{eqn:model_moebm}
            &L_\text{MOE-EBM}({\theta}) = E_{q_{\Phi}(z|X)} \log p_{\beta}(X|z) -\text{KL}\left[q_{\Phi}(z|X) \parallel p_{\alpha}(z) \right]\\
            &= E_{q_{\Phi}(z|X)}\left[\log p_{\beta}(X|z) -\log \frac{q_{\Phi}(z|X)}{p_{\alpha}(z)}\right]\\
            &= E_{q_{\Phi}(z|X)}\left[\log p_{\beta}(X|z) -\log q_{\Phi}(z|X) +\log p_{\alpha}(z)\right]
    \end{aligned}
\end{equation}
}\\
\noindent \textbf{VAE Learning}
By taking derivative of Eqn. \ref{eqn:model_moebm}, we can obtain gradient with respect to ${\theta}$ as shown in Eqn. \ref{eqn:obj_vaebm}, where ${\theta} = ({\beta, \phi})$. By replacing $p_{\alpha}(z)$ with the equation in Eqn. \ref{eqn:model_ebm}, we obtain a refined objective that includes the ELBO and an additional energy term $f_{\alpha}(z)$. When training the VAE part, we consider the term $Z({\alpha})$ as constant since it does not involve sampling from the expectation of $q_{\Phi}(z|X)$.
{
\footnotesize
\begin{equation}
    \begin{aligned}\label{eqn:obj_vaebm}
        &L^{'}_\text{MOE-EBM}({\theta})\\
        &= - E_{q_{\Phi}(z|X)}\left[\frac{\partial}{\partial {\theta}}(\log p_{\beta}(X|z) + \log q_{\Phi}(z|X) - \log p_{\alpha}(z)) \right] \\
        &= \underbrace{- E_{q_{\Phi}(z|X)} \left[\frac{\partial}{\partial {\theta}}(\log p_{\beta}(X|z) + \log \frac{q_{\Phi}(z|X)}{p(z)})\right]}_\text{ELBO}\\
        &+ E_{q_{\Phi}(z|X)} \left[\frac{\partial}{\partial {\theta}} f_{\alpha}(z) \right]
    \end{aligned}
\end{equation}
}

\noindent \textbf{EBM Learning}
For the EBM prior model, as shown in Eqn. \ref{eqn:model_ebm}, we have
{
\small
\begin{equation}
    \begin{aligned}\label{eqn:log_ebm}
        \log p_{\alpha}(z) = f_{\alpha}(z) - \log Z({\alpha}) + \log p(z)
    \end{aligned}
\end{equation}
}

{\noindent  where the derivative of Eqn. \ref{eqn:log_ebm} is as follows:}
{
\small
\begin{equation}
    \begin{aligned}\label{eqn:mle_ebm}
        \frac{\partial}{\partial \alpha}\log p_{\alpha}(z) 
        &=\frac{\partial}{\partial \alpha}f_{\alpha}(z) - E_{p_{\alpha}(z)}\frac{\partial}{\partial \alpha}f_{\alpha}(z)
    \end{aligned}
\end{equation}
}

{\noindent According to Eqn. \ref{eqn:model_moebm} and Eqn. \ref{eqn:mle_ebm}, the learning gradient for ${\alpha}$ is as follows:}
{
\small
\begin{equation}
    \begin{aligned}\label{eqn:learn_ebm}
        &L^{'}_\text{MOE-EBM}({\alpha})=  E_{q_{\phi}(z|X)}\frac{\partial}{\partial {\alpha}}\log p_{\alpha}(z)\\
        &= E_{q_{\Phi}(z|X)}\frac{\partial}{\partial {\alpha}}f_{\alpha}(z) - E_{p_{\alpha}(z)}\frac{\partial}{\partial {\alpha}}f_{\alpha}(z)
    \end{aligned}
\end{equation}
}

\noindent \textbf{MCMC Inference with Langevin Dynamics}
From Eqn. \ref{eqn:learn_ebm}, we notice that learning EBM requires sampling $z$ from two expectations: $E_{q_{\Phi}(z|X)}$ and $E_{p_{\alpha}(z)}$, which can be achieved through MCMC sampling, such as Langevin dynamics (LD), as shown in Eqn. \ref{eqn:ld_sample}.
{
\small
\begin{equation}
\begin{aligned}\label{eqn:ld_sample}
&z_{\tau+1} = z_{\tau} - \frac{s^2}{2}\frac{\partial}{\partial z}[\log {\pi}(z_{\tau})] + s\cdot\epsilon_{\tau}
\end{aligned}
\end{equation}
}

{\noindent Where $s$ is the step size, $\epsilon_{\tau}$ represents Gaussian noise ($\epsilon_{\tau} \sim \mathcal{N}(0,I_d)$), and $\tau$ is the time step in LD. When sampling from EBM prior, ${\pi}(z_{\tau}) = p_{\alpha}(z)$ where $p_{\alpha}(z)$ is initialized from a simple reference distribution. In this work, we use Laplacian distribution as initialization. When sampling from $q_{\Phi}(z|X)$, ${\pi}(z_{\tau}) = q_{\Phi}(z|X)$ where $q_{\Phi}(z|X)$ is initialized from a variational inferred posterior. $M_{\Phi}^{z}(\cdot)$ denotes the Markov transition kernel of finite step LD that samples $z$ from $q_{\Phi}(z|X)$, indicating the marginal distribution of $z$ obtained by running $M_{\Phi}^{z}q_{\Phi}(z|X) = \int_{z^\prime} M_{\Phi}^{z}(z^\prime)q_{\Phi}(z^\prime|X)dz^\prime$ initialized from $q_{\Phi}(z|X)$.}
{\noindent The improved version of learning EBM with MCMC sampling on both the prior and posterior has the following refined format:}
{
\small
\begin{equation}
    \begin{aligned}
        &L^{'}_\text{MOE-EBM}({\alpha})\\
        &= E_{M_{\Phi}^{z}q_{\Phi}(z|X)}\frac{\partial}{\partial {\alpha}}f_{\alpha}(z) - E_{p_{\alpha}(z)}\frac{\partial}{\partial {\alpha}}f_{\alpha}(z)
    \end{aligned}
\end{equation}
}

{\noindent Because we initialize $p_{\alpha}(z)$ from a non-informative $\text{Laplace}(0, I_d)$, but initialize $M_{\Phi}^{z}q_{\Phi}(z|X)$ from a relatively informative variational inferred posterior $q_{\Phi}(z|X)$, in LD for both the prior and posterior, we set different time steps ${\tau}$ and step numbers $s$ to better learn EBM.}\\
\noindent \textbf{MOE with Modality Prior}
To validate our proposed MOE-EBM and compare it with recent MOE-based multimodal generative baselines, we adopt the recent variants of MOE \cite{palumbo2023mmvae+}, which model both shared and modality-specific priors in separate latent subspaces. The detailed design of such a latent space can be referred to in \cite{palumbo2023mmvae+}. We test our proposed model with this latent subspace to investigate its effectiveness within the MOE variant framework.

%% file: sec/4_exp.tex
\section{Experiment}
\subsection{Dataset and Experiment Settings}
To evaluate our model, we use PolyMNIST \cite{sutter2021generalized} to numerically and visually assess the effectiveness of the EBM prior with MCMC inference. A detailed description of PolyMNIST can be found in \cite{sutter2021generalized}. Quantitatively, we measure generative coherence \cite{shi2019variational} to investigate consistency in generations. Additionally, we assess perceptual performance using FID. We test our model on MOE with a modality-specific prior: MMVAE$+$ \cite{palumbo2023mmvae+}. Our results are also compared to other baselines built within the MOE framework, including MMVAE \cite{shi2019variational}, mmJSD \cite{sutter2020multimodal}, and MoPoE \cite{sutter2021generalized}.

\subsection{EBM Prior with MCMC Inference}
\begin{figure}[ht]
     \centering
        \includegraphics[width=0.48\textwidth]{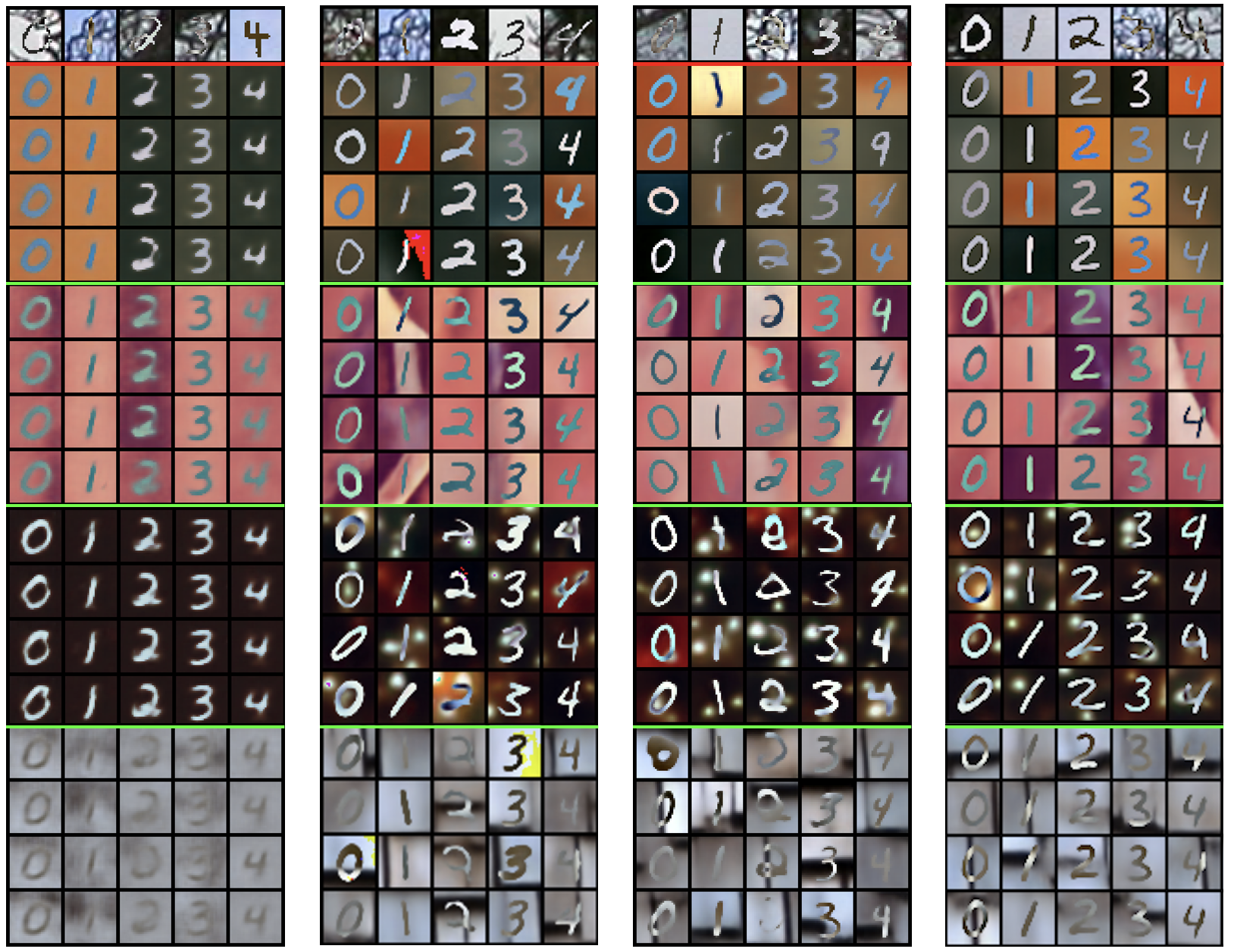}
        \label{fig:moe_based}
    \caption{Cross Generation: from right to left are EBM on MMVAE(MOE), MMVAE$+$, EBM on MMVAE$+$, MOE-EBM on MMVAE$+$}
    \label{fig:cross_visualize}
\end{figure}

Figures \ref{tab:coh_fid_result_poly} and \ref{fig:cross_visualize} show joint and cross generation across different frameworks, with digits highlighted as the shared information among modalities. Both joint and cross-generation demonstrate improvements through visual comparisons with other MOE-based multimodal generative models. Quantitative comparisons in Table \ref{tab:coh_fid_result_poly} further validate the effectiveness of our proposed MOE-EBM.


\begin{table*}[ht]
\centering
\begin{tabular}{c|c|c|c|c}
    \hline
    Model 	& Joint Coherence ($\uparrow$) & Cross Coherence ($\uparrow$) & Joint FID ($\downarrow$) & Cross FID ($\downarrow$)\\
    \hline
    MMVAE*  &0.232 &0.844  &164.71 &150.83 \\
    mmJSD* &0.060 &0.778 &180.55 & 222.09 \\
    MoPoE*  &0.141 &0.720 &107.11 &178.27   \\
    MMVAE+* &0.344 &0.869 &\textbf{96.01} &92.81  \\
    \hline  \hline 
    Ours(MOE-EBM: pre LD) &NA  &\textbf{0.885} &NA & 94.72  \\
    \hline
    Ours(MOE-EBM: post LD) &\textbf{0.574}  &\textbf{0.943} &98.23 & \textbf{90.32}  \\
    \hline
\end{tabular}\\
\caption{Generation Coherence and FID : We present generation results of our proposed MOE-EBM with before LD (variational inference, denoted as pre LD) and after LD (MCMC inference , denoted as post LD), and compare with other MOE-based multimodality generative models ( $*$ are results referred from \cite{palumbo2023mmvae+}.)}
\label{tab:coh_fid_result_poly}
\end{table*}

\begin{figure}[ht]
     \centering
     \begin{subfigure}[b]{0.235\textwidth}
        \includegraphics[width=\textwidth]{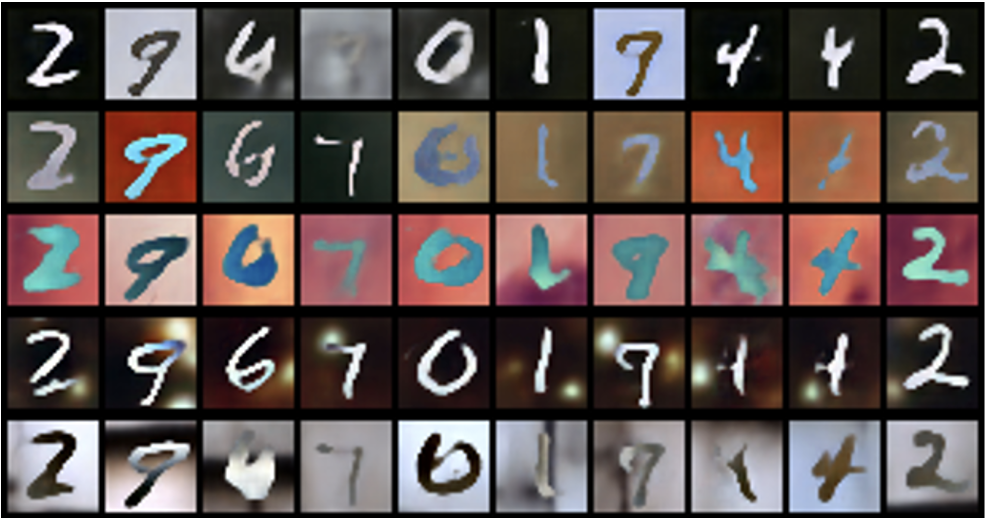}
        \caption{MOE-EBM on MMVAE$+$}
        \label{fig:ebm-mcmc_iclr23}
    \end{subfigure}   
    \hfill
    \begin{subfigure}[b]{0.235\textwidth}
        \includegraphics[width=\textwidth]{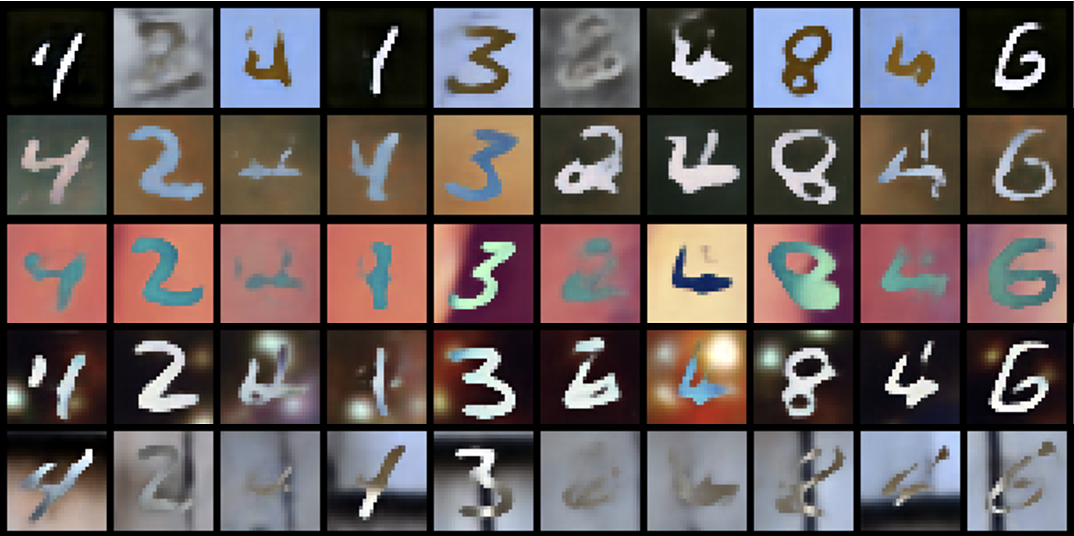}
        \caption{EBM on MMVAE$+$}
        \label{fig:ebm_prior_iclr23}     
    \end{subfigure}
        \hfill
    \begin{subfigure}[b]{0.235\textwidth}
        \includegraphics[width=\textwidth]{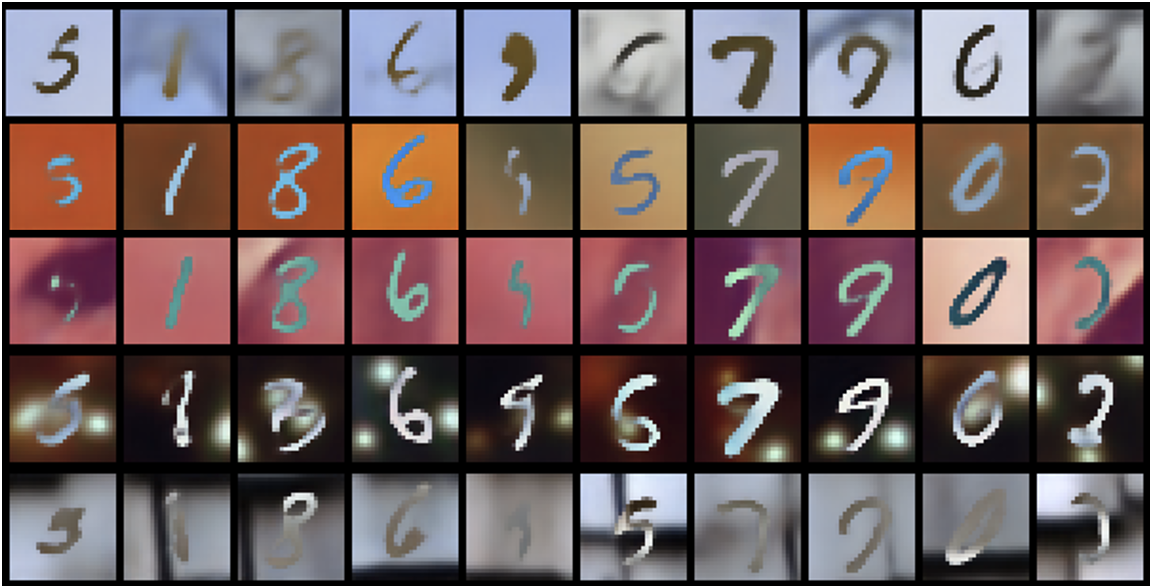}
        \caption{MMVAE$+$ }
        \label{fig:uni_prior_iclr23}     
    \end{subfigure}
        \hfill
    \begin{subfigure}[b]{0.235\textwidth}
        \includegraphics[width=\textwidth]{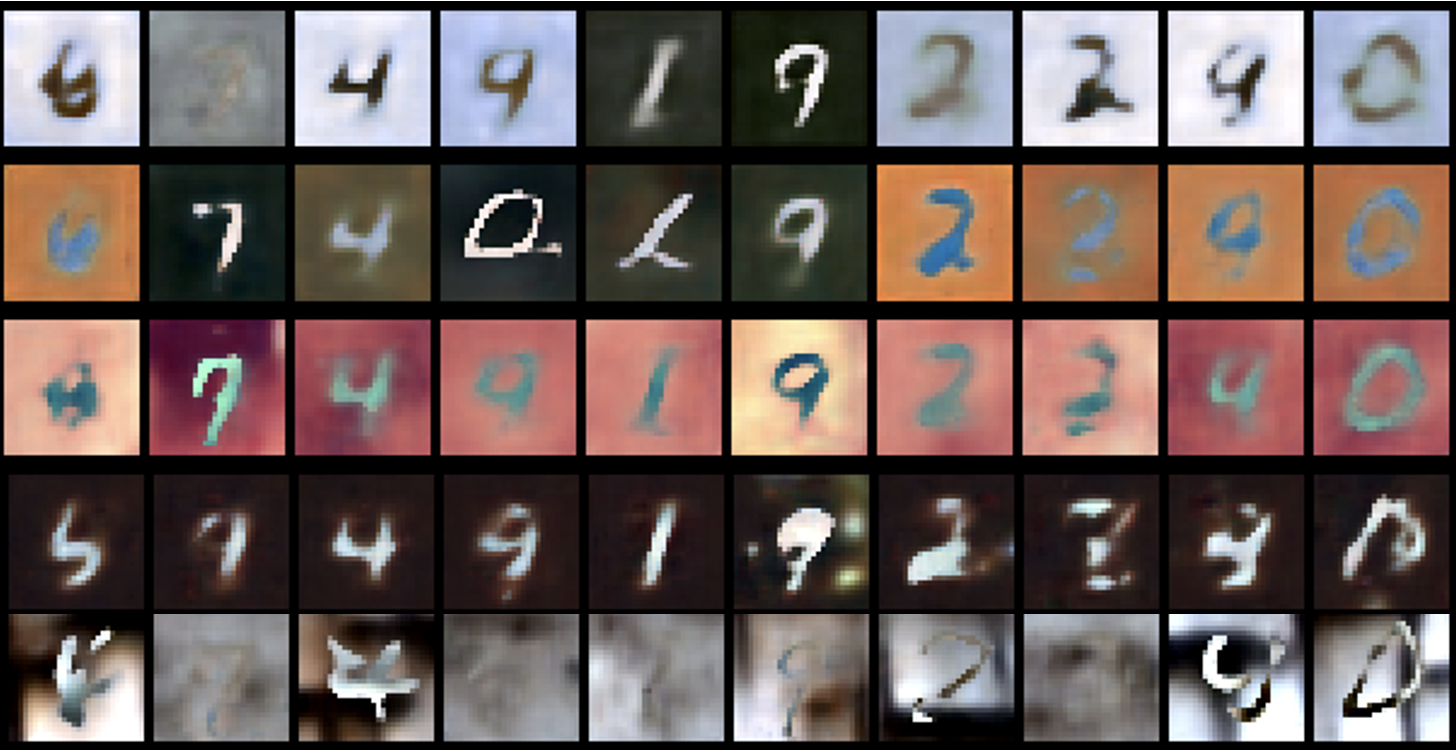}
        \caption{MMVAE(MOE)}
        \label{fig:uni_prior_base}     
    \end{subfigure}   
    \caption{Joint Generation}
    \label{fig:joint_visualize}
\end{figure}
\vspace{-3mm}
\subsection{Generation Comparison between Variational Inference and MCMC Inference}
\begin{figure}[ht]
    \centering
    \begin{subfigure}[b]{0.48\textwidth}
        \includegraphics[width=\textwidth]{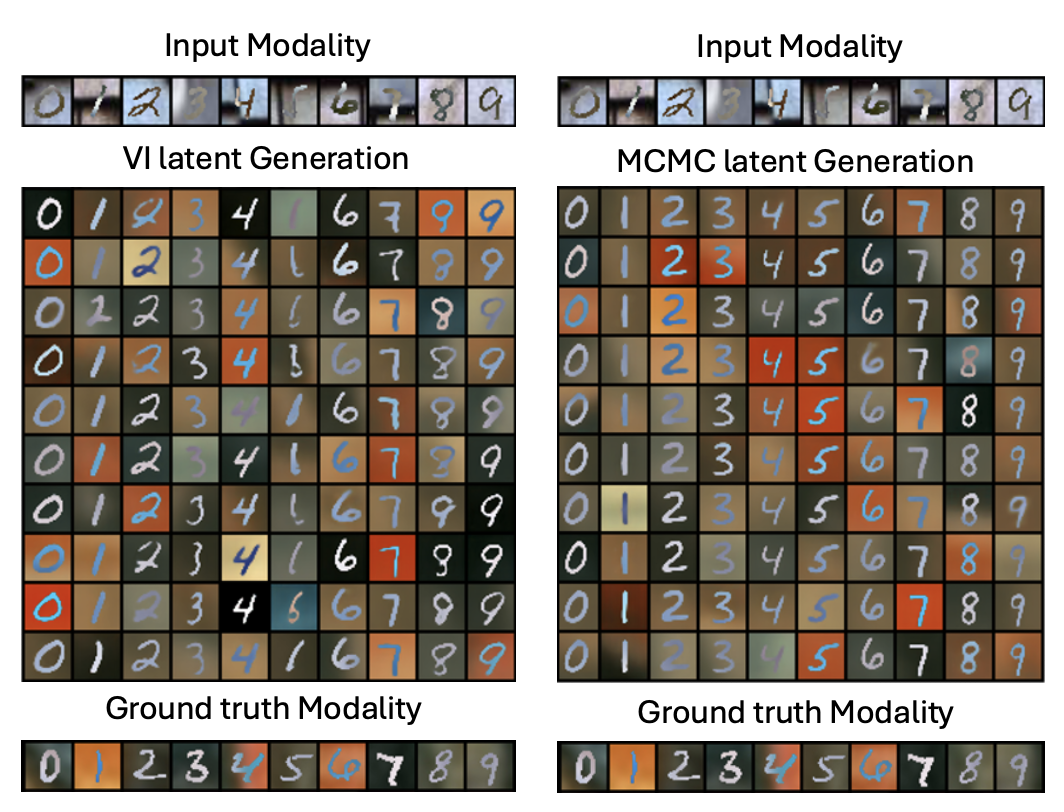}
        \caption{Variational Inference vs. MCMC Inference Generation}
        \label{fig:vi_mcmc_gen}
    \end{subfigure}
    \begin{subfigure}[b]{0.45\textwidth}
        \centering
        \includegraphics[width=\textwidth]{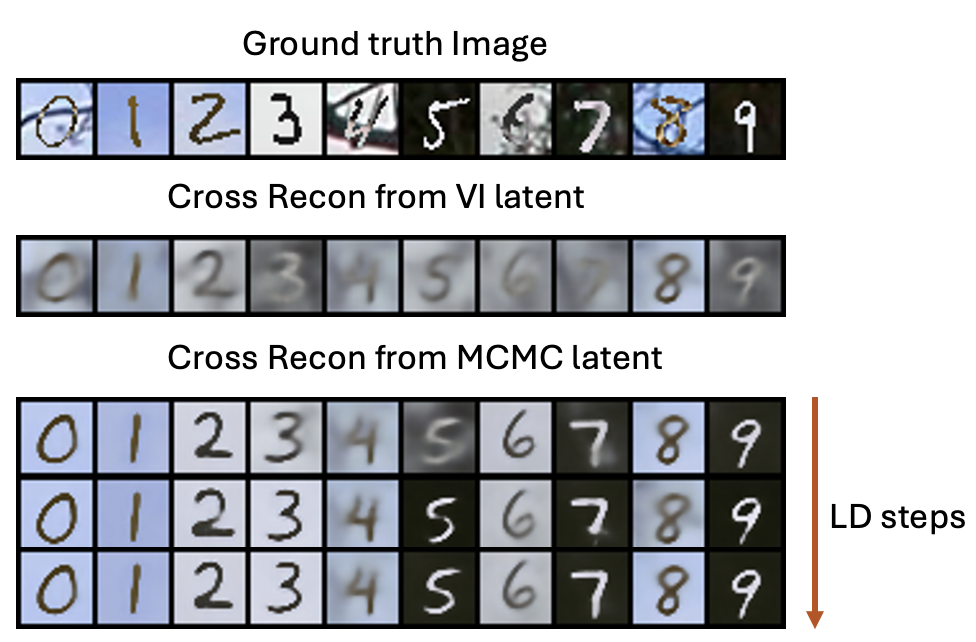}
        \caption{Generation in Markov transition with LD}
        \label{fig:jointebm_prior}
    \end{subfigure}
    \caption{(a) Comparative visualization of generation quality before and after LD refinement. (b) Generation improvement during Markov transitions using LD.}
    \label{fig:ld_visual}
\end{figure}
MCMC inference learned through EBM can be closely to the true posterior compare with variational inferred posterior. We explore the generation quality before and after LD in Figure \ref{fig:vi_mcmc_gen} and visualize the changes in generation quality during LD refinement in Figure \ref{fig:ld_visual}. To quantitatively validate that MCMC inference closely approximates the true posterior, we present the generation coherence before and after LD in Table \ref{tab:coh_fid_result_poly}.

%% file: sec/5_ablation.tex
\section{Ablation Studies}
To investigate the effectiveness of an EBM prior with MCMC inference in multimodal generation, we conduct two ablation studies: one incorporating an EBM prior with MOE, and the other incorporating an EBM prior with MOE variants, learning with a modality-specific prior. Notably, neither ablation involved MCMC inference. We present the results for each setting in comparison with our MOE-EBM framework in Tables \ref{tab:ablation_coh_result_poly} and \ref{tab:ablation_fid_result_poly}. We observe that using only the EBM prior, generation coherence showed non-trivial improvements compared to the corresponding baselines. This indicates that the EBM prior can better capture shared information in the complex multimodal data space. Furthermore, MCMC inference directly benefits cross-modal generation, as validated by the ablation results.
\begin{table}[ht]
\centering
\begin{tabular}{c|c|c}
    \hline
    Model 	& Joint Coh($\uparrow$) & Cross Coh ($\uparrow$) \\
    \hline
    EBM-MMVAE &0.340  &0.856   \\
    \hline
    EBM-MMVAE$+$ &0.531  &0.877   \\
    \hline
    \hline
    MOE-EBM &\textbf{0.574}  &\textbf{0.943}\\
    \hline
\end{tabular}\\
\caption{Ablation Results: Generation Coherence (Coh refer to Coherence due to space limitation) }
\label{tab:ablation_coh_result_poly}
\end{table}
\vspace{-6mm}
\begin{table}[ht]
\centering
\begin{tabular}{c|c|c}
    \hline
    Model &Joint FID ($\downarrow$) & Cross FID ($\downarrow$)\\
    \hline
    EBM-MMVAE  &129.66 &152.01  \\
    \hline
    EBM-MMVAE$+$ &100.65 & 95.37  \\
    \hline
    \hline
    MOE-EBM  &\textbf{98.23} & \textbf{90.32}  \\
    \hline
\end{tabular}\\
\caption{Ablation Results: FID}
\label{tab:ablation_fid_result_poly}
\end{table}

%% file: sec/5_conclusion.tex
\section{Future work}
We plan to focus on two main avenues for future research. First, we will explore additional multimodal datasets, particularly those with high-resolution real images. Besides assessing generative coherence and perceptual performance, we aim to evaluate our model on various analytical tasks, including latent space analysis and mutual information analysis.
